\documentclass[runningheads]{llncs}

\usepackage{eccv}

\usepackage{eccvabbrv}

\usepackage{graphicx}
\usepackage{amsmath}
        
\usepackage{booktabs}
\usepackage{multirow}

\usepackage[normalem]{ulem}
\useunder{\uline}{\ul}{}

\usepackage[accsupp]{axessibility}  

\usepackage[breaklinks,colorlinks]{hyperref}

\usepackage{orcidlink}

\begin{document}

\title{TalkingGaussian: Structure-Persistent 3D Talking Head Synthesis via Gaussian Splatting} 

\titlerunning{TalkingGaussian}

\author{Jiahe Li\inst{1} \and
Jiawei Zhang\inst{1} \and
Xiao Bai\inst{1}\thanks{Corresponding author: Xiao Bai (baixiao@buaa.edu.cn).} \and
Jin Zheng\inst{1} \and
Xin Ning\inst{2} \and
Jun Zhou\inst{3} \and
Lin Gu\inst{4,5} \\
}

\authorrunning{J. Li et al.}

\institute{School of Computer Science and Engineering, State Key Laboratory of \\ Complex \& Critical Software Environment, Jiangxi Research Institute, \\ Beihang University \and
Institute of Semiconductors, Chinese Academy of Sciences \and
School of Information and Communication Technology, Griffith University \and
RIKEN AIP \and 
The University of Tokyo}

\maketitle

\begin{figure}
    \centering
    \vspace{-5mm}
    \includegraphics[width=1\linewidth]{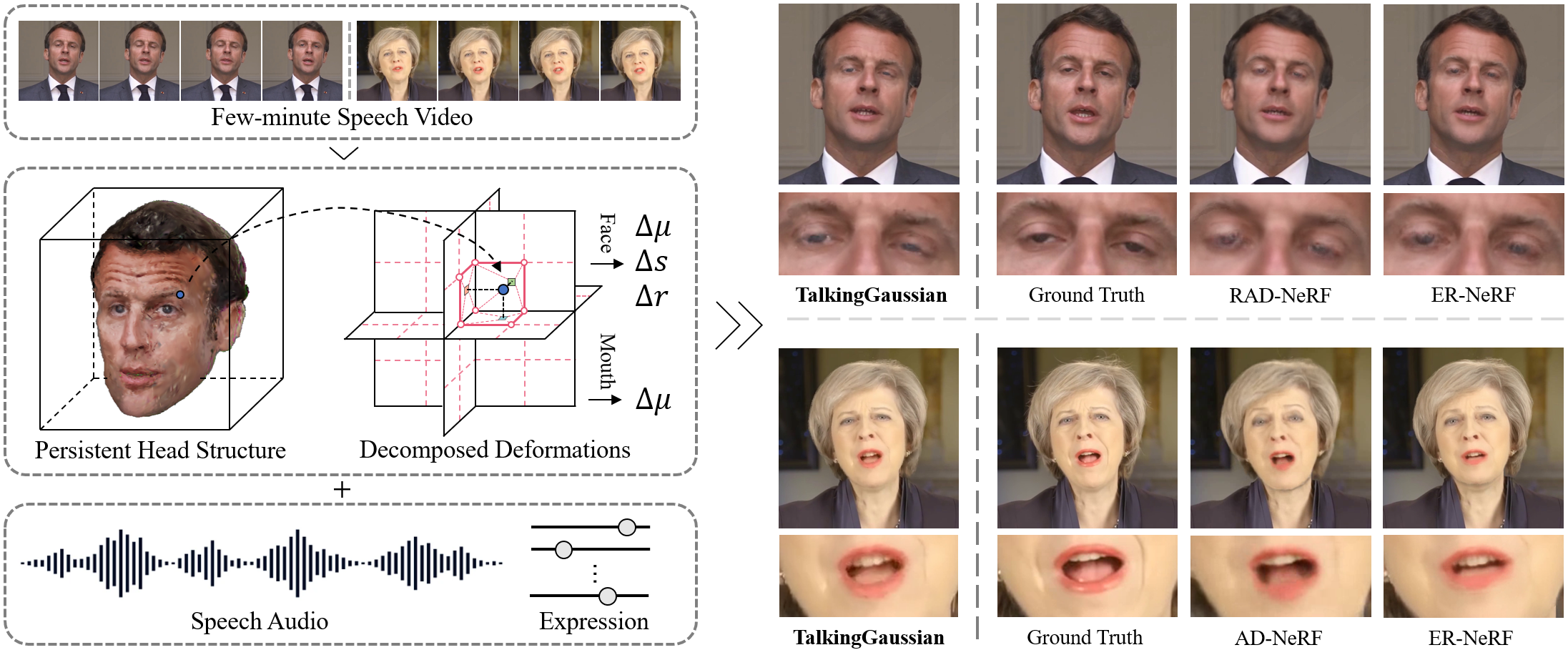}
    \vspace{-5mm}
    \caption{Inaccurate predictions of the rapidly changing appearance often produce distorted facial features in previous NeRF-based methods. By keeping a persistent head structure and predicting deformation to represent facial motion, our TalkingGaussian outperforms previous methods in synthesizing more precise and clear talking heads.}
    \label{fig:teaser}
    \vspace{-10mm}
\end{figure}

\begin{abstract}
Radiance fields have demonstrated impressive performance in synthesizing lifelike 3D talking heads. However, due to the difficulty in fitting steep appearance changes, the prevailing paradigm that presents facial motions by directly modifying point appearance may lead to distortions in dynamic regions. To tackle this challenge, we introduce TalkingGaussian, a deformation-based radiance fields framework for high-fidelity talking head synthesis. Leveraging the point-based Gaussian Splatting, facial motions can be represented in our method by applying smooth and continuous deformations to persistent Gaussian primitives, without requiring to learn the difficult appearance change like previous methods. Due to this simplification, precise facial motions can be synthesized while keeping a highly intact facial feature. Under such a deformation paradigm, we further identify a face-mouth motion inconsistency that would affect the learning of detailed speaking motions. To address this conflict, we decompose the model into two branches separately for the face and inside mouth areas, therefore simplifying the learning tasks to help reconstruct more accurate motion and structure of the mouth region. 
Extensive experiments demonstrate that our method renders high-quality lip-synchronized talking head videos, with better facial fidelity and higher efficiency compared with previous methods.

\end{abstract}

\section{Introduction}

Synthesizing audio-driven talking head videos is valuable to a wide range of digital applications such as virtual reality, film-making, and human-computer interaction. Recently,  radiance fields like Neural Radiance Fields (NeRF) \cite{mildenhall2021nerf} have been adopted by many methods \cite{guo2021ad, tang2022rad, li2023efficient, ye2023geneface, peng2023synctalk, shen2022dfrf, Chatziagapi2023lipnerf} to improve the stability of 3D head structure while providing photo-realistic rendering, which has achieved great success in synthesizing high-fidelity talking head videos.

Most of these NeRF-based approaches \cite{guo2021ad, tang2022rad, li2023efficient, ye2023geneface, peng2023synctalk} synthesize different face motions by directly modifying color and density with neural networks, predicting a temporary condition-dependent appearance for each spatial point in the radiance fields whenever receiving a condition feature. This appearance-modification paradigm enables previous methods to achieve dynamic lip-audio synchronization in a fixed space representation. However, since even neighbor regions can also show significantly different colors and various structures on a human face, it's challenging for these continuous and smooth neural fields to accurately fit the rapidly changing appearance to represent facial motions, which may lead to some heavy distortions on the facial features like a messy mouth and transparent eyelids, as shown in Fig. \ref{fig:teaser}.

In this paper, we propose TalkingGaussian, a deformation-based talking head synthesis framework, that attempts to utilize the recent 3D Gaussian Splatting (3DGS) \cite{kerbl2023gaussian} to address the facial distortion problem in existing radiance-fields-based methods. 
The core idea of our method is to represent complex and fine-grained facial motions with several individual smooth deformations to simplify the learning task. 
To achieve this goal, we first obtain a persistent head structure that keeps an unchangeable appearance and stable geometry with 3DGS. Then, motions can be precisely represented just by the deformation applied to the head structure, therefore eliminating distortions produced from inaccurately predicted appearance, and leading to better facial fidelity while synthesizing high-quality talking heads.

Specifically, we represent the dynamic talking head with a 3DGS-based Deformable Gaussian Field, consisting of a static Persistent Gaussian Field and a neural Grid-based Motion Field to decouple the persistent head structure and dynamic facial motions. 
Unlike previous continuous neural-based backbones \cite{mildenhall2021nerf, muller2022instant, li2023efficient}, 3DGS provides an explicit space representation by a definite set of Gaussian primitives, enabling us to obtain a more stable head structure and accurate control of spatial points. 
Based on this, we apply a point-wise deformation, which changes the position and shape of each primitive while persisting its color and opacity, to represent facial motions via the motion fields. Then the deformed primitives are input into the 3DGS rasterizer to render the target images. To facilitate the smooth learning for a target facial motion, we introduce an incremental sampling strategy that utilizes face action priors to schedule the optimization process of deformation.

In the Deformable Gaussian Fields, we further decompose the entire head as a face branch and an inside mouth branch to solve the motion inconsistency between these two regions, which hugely improves the synthesis quality both in static structure and dynamic performance. Since the motions of the face and inside mouth are not related totally in tight and may be much different sometimes, it is hard to accurately represent these delicate but conflicted motions with just one single motion field. To simplify the learning of both these two distinct motions, we divide these two regions in 2D input images with a semantic mask, and build two model branches to represent them individually. As the motion in each branch has been simplified to become smooth, our method can achieve better visual-audio synchronization and reconstruct a more accurate mouth structure.

The main contributions of our paper are summarized as follows:
\begin{itemize}
    \item We present a novel deformation-based framework that synthesizes talking heads by applying deformations to a persistent head structure, to escape an inherent facial distortion problem from the inaccurate prediction of changing appearance, enabling the generating of precise and intact facial details.
    \item We propose a Face-Mouth Decomposition module to facilitate motion modeling via decomposing conflicted learning tasks for deformation, therefore providing accurate mouth reconstruction and lip synchronization.
    \item Extensive experiments show that the proposed TalkingGaussian renders realistic lip-synchronized talking head videos with high visual quality, generalization ability, and efficiency, outperforming state-of-the-art methods on both objective evaluation and human judgment.
\end{itemize}


\section{Related Work}

\vspace{4pt}\noindent\textbf{Talking Head Synthesis. }
Driving talking heads by arbitrary input audio is an active research topic, aiming to reenact the specific person to generate highly audio-visual consistent videos. 
Early methods based on 2D generative models synthesize audio-synchronized lip motions for a given facial image \cite{prajwal2020wav2lip, ezzat2002trainable, jamaludin2019you, chen2019hierarchical, wiles2018x2face}. Later advancements \cite{thies2020nvp, wang2020mead, zhang2021facial, lu2021lsp} incorporate intermediate representations like facial landmarks and morphable models for better control, but suffer from errors and information loss during the intermediate estimation. Due to the lack of an explicit 3D structure, these 2D-based methods are short in keeping the naturalness and consistency when the head pose changes.

Recently, Neural Radiance Fields (NeRF)~\cite{mildenhall2021nerf} has been introduced as a 3D representation of the talking head structure, providing photorealistic rendering and personalized talking style via person-specific training. Earlier NeRF-based works \cite{guo2021ad, shen2022dfrf, liu2022semantic} suffer from the expensive cost of vanilla NeRF. Successfully driving efficient neural fields \cite{muller2022instant, chan2022efficient} with audio, RAD-NeRF \cite{tang2022rad} and ER-NeRF \cite{li2023efficient} have gained tremendous improvements in both visual quality and efficiency. To improve the generalizability of cross-domain audio inputs, GeneFace \cite{ye2023geneface} and SyncTalk \cite{peng2023synctalk} pre-train the audio encoder with large audio-visual datasets. However, most of these methods represent facial motions by changing the appearance of each sampling point, which burdens the network with learning the jumping and unsmooth appearance changes, resulting in distorted facial features. Although some works \cite{shen2022dfrf, li2023hide, ye2024real3d} have introduced a pre-trained deformable fields module for few-shot settings, the lack of fine-grained point control and precise head structure brings drawbacks in static and dynamic quality. Instead, utilizing 3DGS to maintain an accurate head structure, our method simplifies the learning difficulty of facial motions with a pure deformation representation, therefore improving facial fidelity and lip-synchronization.

\vspace{4pt}\noindent\textbf{Deformation in Radiance Fields. } Deformation has been widely applied in radiance fields to synthesize dynamic novel views. Some NeRF methods \cite{pumarola2021dnerf, park2021nerfies, park2021hypernerf, song2022nerfplayer, fang2022tineuvox} use a static canonical radiance field to capture geometry and appearance and a time-dependent deformation field for dynamics. These methods predict an offset referring to the sampling position, which is opposite to the motion path and would bring extra difficulties in fitting. To solve this problem, \cite{guo2023forward} use a deformation that directly warps the canonical fields to represent dynamics. However, this method is costly since the spatial points cannot be accurately and stably controlled in its grid-based NeRF representation.

More recently, 3D Gaussian Splatting \cite{kerbl2023gaussian} introduces an explicit point-based representation for radiance fields, where deformation can be easily applied to a definite set of Gaussian primitives to directly warp the canonical fields. Based on this idea, considerable dynamic 3DGS works \cite{luiten2023dynamic3dgs, yang2023deformable3d, wu20234dgaussian, lin2023aussian-flow, kratimenos2023dynmf} get significant improvements in visual quality and efficiency for dynamic novel views synthesis. However, these methods only aim to remember the fixed motion at each time stamp, insufficient to represent various fine-grained motions driven by conditions, especially on the mouth. Despite some attempts conducted to reconstruct the human head \cite{qian2023gaussianavatars, xu2023gaussianheadavatar, chen2023monogaussianavatar, wang2024gaussianhead} driven by parametrized facial models, the mapping from audio to these parameters is not easy to learn and would cause information loss, and thus they can not be easily transferred to our audio-driven task. In this paper, we introduce deformable Gaussian fields with an incremental sampling strategy to facilitate learning multiple complex facial motions from a monocular speech video via pure deformation, and decomposite inconsistent motions of the face and inside mouth areas to improve the quality of delicate talking motions.


\section{Method}

\subsection{Preliminaries and Problem Setting}

\vspace{0pt}\noindent\textbf{3D Gaussian Splatting. } 
3D Gaussian splatting (3DGS) \cite{kerbl2023gaussian} represents 3D information with a set of 3D Gaussians. It computes pixel-wise color $\mathcal{C}$ with a set of 3D Gaussian primitives $\theta$ and the camera model information at the observing view. Specifically, a Gaussian primitive can be described with a center $\mu \in \mathbb{R}^3$, a scaling factor $s \in \mathbb{R}^3$, and a rotation quaternion $q \in \mathbb{R}^4$. For rendering purposes, each Gaussian primitive also retains an opacity value $\alpha \in \mathbb{R}$ and a $Z$-dimensional color feature $f \in \mathbb{R}^Z$. Thus, the $i$-th Gaussian primitive $\mathcal{G}_i$ keeps a set of parameters $\theta_i = \{\mu_i, s_i, q_i, \alpha_i, f_i\}$. Its basis function is in the form of:
\begin{equation}
    \setlength{\abovedisplayskip}{4pt}
    \setlength{\belowdisplayskip}{4pt}
    \mathcal{G}_i(\mathbf{x}) = e^{-\frac{1}{2}(\mathbf{x-\mu_i})^T\Sigma_i^{-1}(\mathbf{x-\mu_i})},
\end{equation}
where the covariance matrix $\Sigma$ can be calculated from $s$ and $q$.

During the point-based rendering, a rasterizer would gather $N$ Gaussians following the camera model to compute the color $\mathcal{C}$ of pixel $\mathbf{x}_p$, with the decoded color $c$ of feature $f$ and the projected opacity $\widetilde{\alpha}$ calculated by their projected 2D Gaussians $\mathcal{G}^{proj}$ on image plane:
\begin{equation}
    \setlength{\abovedisplayskip}{3pt}
    \setlength{\belowdisplayskip}{3pt}
    \mathcal{C}(\mathbf{x}_p) = \sum_{i \in N}{c_i\widetilde{\alpha}_i\prod_{j=1}^{i-1}(1-\widetilde{\alpha}_j)}, \quad \widetilde{\alpha}_i = \alpha_i \mathcal{G}^{proj}_i(\mathbf{x}_p).
\end{equation}
Similarly, the opacity $\mathcal{A} \in [0,1]$ of pixel $\mathbf{x}_p$ can be given:
\begin{equation}
    \setlength{\abovedisplayskip}{3pt}
    \setlength{\belowdisplayskip}{3pt}
    \mathcal{A}(\mathbf{x}_p) = \sum_{i \in N}{\widetilde{\alpha}_i\prod_{j=1}^{i-1}(1-\widetilde{\alpha}_j)}.
\end{equation}

3DGS optimizes the parameters $\theta$ for all Gaussians through gradient descent under color supervision. During the optimization process, it applies a densification strategy to control the growth of the primitives, while also pruning unnecessary ones. This work inherits these optimization strategies for color supervision.

\vspace{2pt}\noindent\textbf{Problem Setting. } 
In this paper, we aim to present an audio-driven framework based on 3DGS representation for high-fidelity talking head synthesis. Adopting a similar problem setting as NeRF-based works \cite{guo2021ad, liu2022semantic, tang2022rad, li2023efficient}, we take a few-minute speech video with a single person as the training data. A 3DMM model \cite{paysan2009bfm} is utilized to estimate the head pose and therefore to infer the camera pose. To keep aligned with previous works \cite{guo2021ad, shen2022dfrf, liu2022semantic, li2023efficient}, we use a pre-trained DeepSpeed \cite{hannun2014deepspeech} model as the basic audio encoder to get a generalizable audio feature from the raw input speech audio.

\begin{figure}[t]
    \centering
    \includegraphics[width=\linewidth]{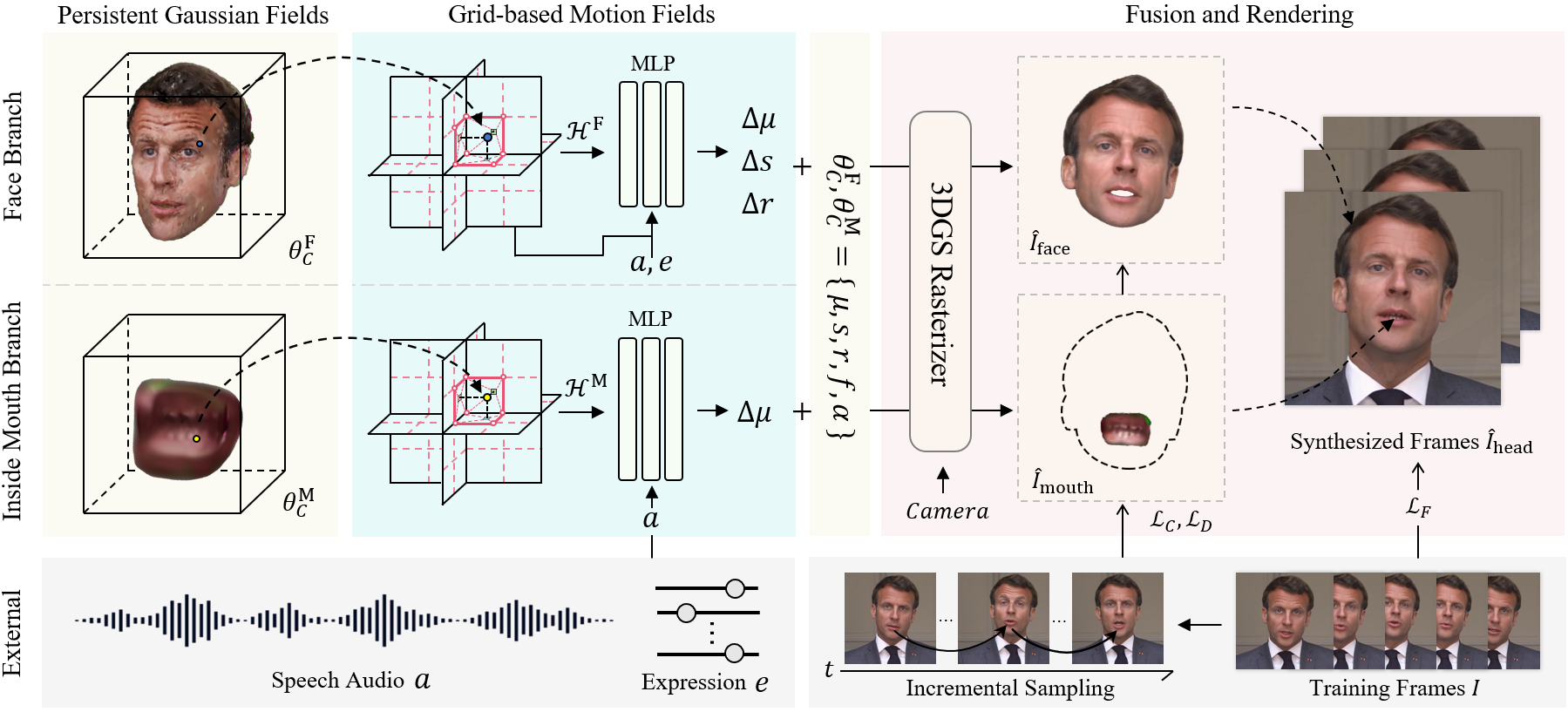}
    \caption{\textbf{Overview of TalkingGaussian.} Learning from the speech video with training frames $I$, TalkingGaussian builds two separate branches to represent the dynamic face and inside mouth areas. Queried by the primitives in Persistent Gaussian Fields with parameters $\theta_C$, a point-wise deformation can be predicted from Grid-based Motion Fields conditioned with audio feature $\boldsymbol{a}$ and upper-face expression $\boldsymbol{e}$. After that, the 3DGS rasterizer renders the deformed 3D Gaussian primitives into 2D images observed from the given camera, which are then fused to synthesize the entire talking head.}
    \label{fig:main}
    \vspace{-4mm}
\end{figure}

\subsection{Deformable Gaussian Fields for Talking Head. \label{sec:fields}}

\begin{figure}[t]
    \centering
    \includegraphics[width=\linewidth]{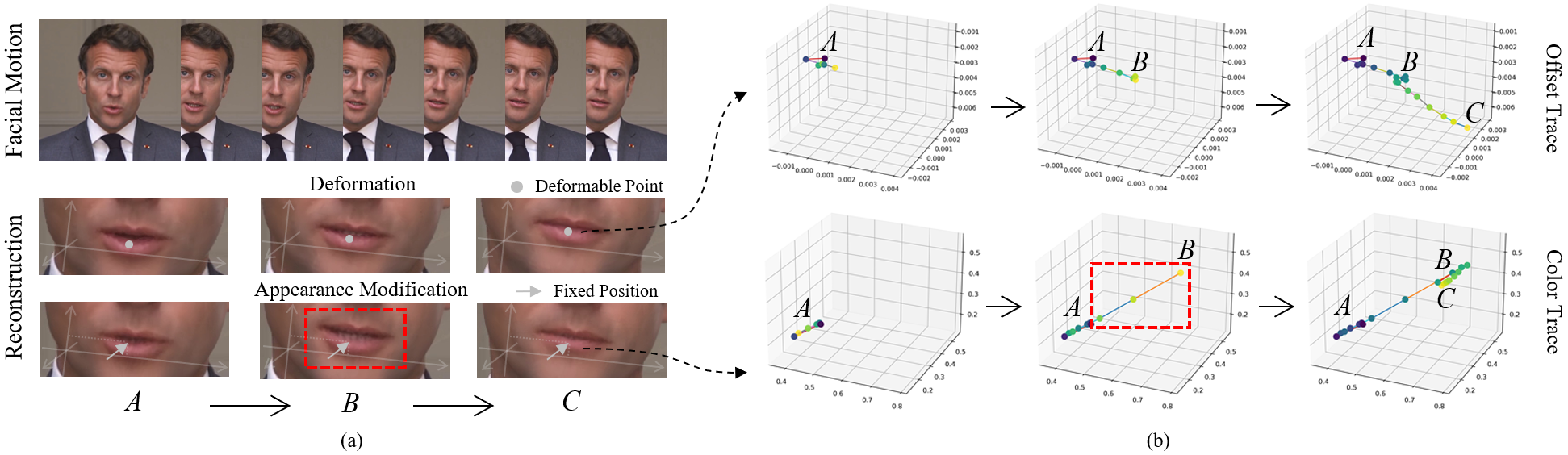}
    \vspace{-7mm}
    \caption{\textbf{(a)} The reconstructed facial motion results represented by deformation and appearance modification. \textbf{(b)} The visualized traces of the changing coordinate offset (deformation) and color in RGB (appearance modification) of two points with the same initial position. During the process, offset changes smoothly and the corresponding results are clear and accurate.  Instead, some sudden changes with a large step length may occur in color, which is difficult to fit and causes a distorted mouth (red box).}
    \label{fig:space}
    \vspace{-3mm}
\end{figure}


Despite previous NeRF-based methods \cite{guo2021ad, shen2022dfrf, tang2022rad, ye2023geneface, li2023efficient, liu2022semantic} have achieved great success in synthesizing high-quality talking heads via generating point-wise appearance, they can still not tackle the problem of generating distorted facial features on dynamic regions. One main reason is the appearance space, including color and density, is jumping and unsmooth, which makes it difficult for the continuous and smooth neural fields to fit. In comparison, deformation is another choice to represent motions with better smoothness and continuity, as shown in Fig. \ref{fig:space}. In this work, we propose to purely use deformation in the Gaussian radiance fields to represent different motions of the talking head in 3D space. In particular, the whole representation is decomposed into Persistent Gaussian Fields and Grid-based Motion Fields, as shown in Fig. \ref{fig:main}. These fields will be further refined for different regions in the next section.

\vspace{4pt}\noindent\textbf{Persistent Gaussian Fields. }
Persistent Gaussian Fields preserve the persistent Gaussian primitive with the canonical parameters $\theta_C = \{\mu, s, q, \alpha, f\}$. Firstly, we initialize this module with the static 3DGS by the speech video frames to get a coarse mean field. Later, it attends a joint optimization with the Grid-based Motion Fields.  

\vspace{4pt}\noindent\textbf{Grid-based Motion Fields. }
Although the primitives in Persistent Gaussian Fields can effectively represent the correct 3D head, a regional position encoding is lacking due to their fully explicit space structure. Considering most facial motions are regionally smooth and continuous, we adopt an efficient and expressive tri-plane hash encoder $\mathcal{H}$ \cite{li2023efficient} for position encoding with an MLP decoder to build Grid-based Motion Fields for a continuous deformation space. 

Specifically, the motion fields aim to represent the facial motion by predicting a point-wise deformation $\delta_i = \{\Delta 
\mu_i, \Delta s_i, \Delta q_i\}$ for each primitive with the input of its center $\mu_i$, which is irrelevant to the color and opacity changing. For the given condition feature set $\mathbf{C}$, the deformation $\delta_i$ can be calculated by:
\begin{equation}
    \delta_i = \text{MLP}(\mathcal{H}(\mu_i) \oplus \mathbf{C}),
\end{equation}
where $\oplus$ denotes concatenation.

Through a 3DGS rasterizer, these two fields are combined to generate deformed Gaussian primitives to render the output image, of which the deformed parameters $\theta_D$ are got from the canonical parameters $\theta_C$ and deformation $\delta$:
\begin{equation}
    \setlength{\abovedisplayskip}{4pt}
    \setlength{\belowdisplayskip}{4pt}
    \label{eq:deform_param}
    \theta_D = \{\mu+\Delta \mu, s+\Delta s, q+\Delta q, \alpha, f\}.
\end{equation}

\vspace{4pt}\noindent\textbf{Optimization with Incremental Sampling. }
While learning the deformation, once the target primitive position is too far from the predicted results, the gradient would vanish and thus the motion fields may fail to be effectively updated. To tackle this problem, we introduce an incremental sampling strategy. Specifically, we first find a valid metric $m$ (e.g. action units \cite{Ekman1978FacialAC} or landmarks) to measure the deformation degree of each target facial motion. Then, at the $k$-th training iteration, we use a sliding window to sample a required training frame at position $j$, of which the motion metric $m_j$ satisfies the condition:
\begin{equation}
    \setlength{\abovedisplayskip}{4pt}
    \setlength{\belowdisplayskip}{4pt}
    m_j \in [B_{lower}+k \times T, B_{upper}+k \times T],
\end{equation}
where $B_{lower}$ and $B_{upper}$ denote the initial lower and upper bound of the sliding window, and $T$ denotes the step length.
This selected training frame can offer sufficient new knowledge for the deformable fields to learn, but would not be too hard.
To avoid catastrophic forgetting, we apply the incremental sampling strategy once every $K$ iterations.

\subsection{Face-Mouth Decomposition}

\begin{figure}[t]
    \centering
    \includegraphics[width=\linewidth]{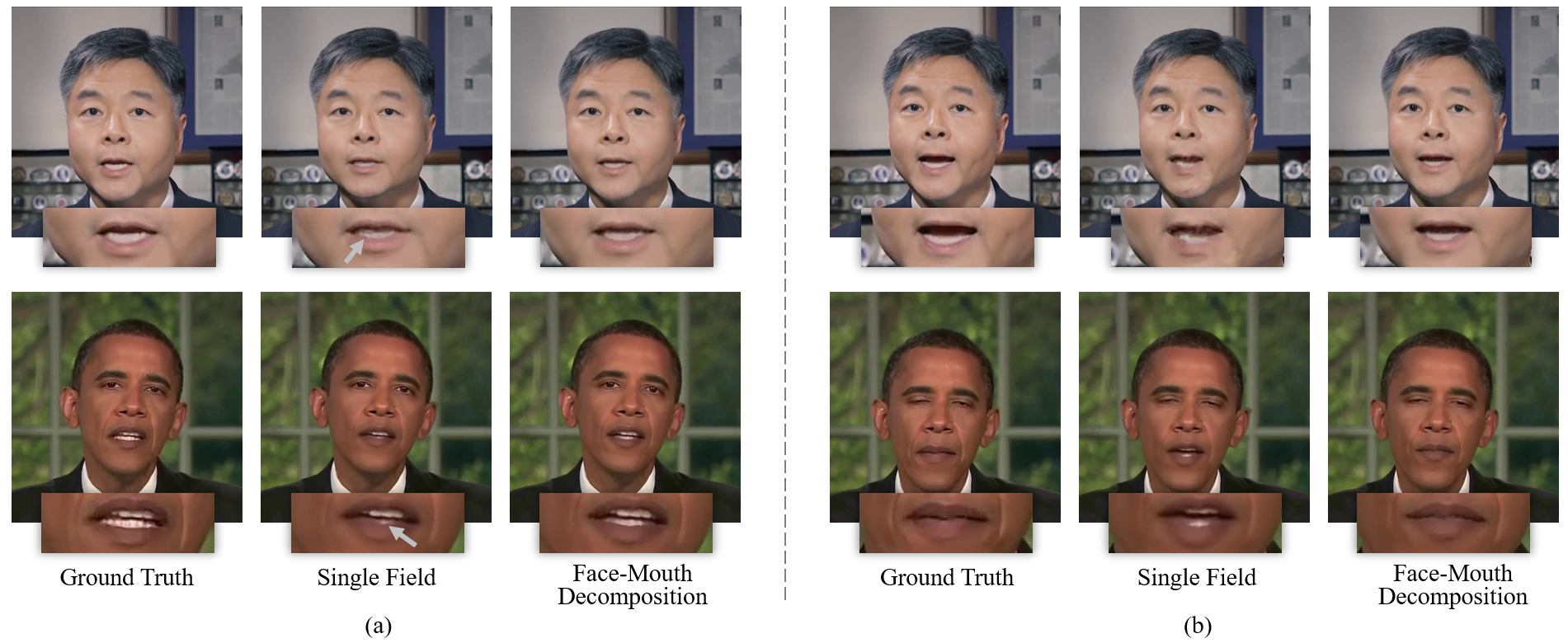}
    \vspace{-6mm}
    \caption{\textbf{(a)} Lips and the inside mouth, especially teeth, are hard to be correctly divided with a single motion field.  \textbf{(b)} This would further affect the learning of the mouth structure and speaking motions, resulting in bad quality. Our Face-Mouth Decomposition can successfully address this problem and render high-fidelity results.}
    \label{fig:decomp}
    \vspace{-4mm}
\end{figure}

Although the Grid-based Motion Fields can predict the point-wise deformation at arbitrary positions due to the continuous and dense 3D space representation, this representation still encounters a granularity problem caused by the motion inconsistency between the face and the inside mouth. Since the inside area of the mouth is spatially too close to the lips but does not always move together, their motions would interfere with each other in a single interpolation-based motion field. This can also further lead to a bad reconstruction quality in static structure as well, as shown in Fig. \ref{fig:decomp}. 

To tackle this problem, we propose decomposing these two regions in 3D space and building two individual branches with separate optimization.
For each training video frame, we first use the off-the-shelf face parsing models to get a semantic mask of the inside mouth region in 2D space \footnote{Additional descriptions and details can be found in the supplementary material. \label{fn: supp}}. Then, we take the masked image of the inside mouth and the remaining surface region (containing the face, hair, and other head parts) to train two separate deformable Gaussian fields as two branches of our framework.

\vspace{4pt}\noindent\textbf{Face Branch. }
The face branch serves as the main part to fit the appearance and motion of the talking head, including all facial motions except the one of the inside mouth. 
In this branch, we adopt a region attention mechanism \cite{li2023efficient} in the Grid-based Motion Fields to facilitate the learning of the conditioned deformation driven by the features of audio $\boldsymbol{a}$ and upper-face expression $\boldsymbol{e}$. To fully decouple these two conditions, the upper-face expression feature $\boldsymbol{e}$ is composed of 7 action units \cite{Ekman1978FacialAC} that are explicitly irrelevant to the mouth. The deformation $\delta_i^\text{F}$ for the $i$-th primitive in the face branch can be predicted by: 
\begin{equation}
    \setlength{\abovedisplayskip}{4pt}
    \setlength{\belowdisplayskip}{4pt}
    \delta_i^\text{F} = \text{MLP}(\mathcal{H}^\text{F}(\mu_i) \oplus \boldsymbol{a}_{r, i} \oplus \boldsymbol{e}_{r, i}),
\end{equation}
where $\boldsymbol{a}_{r, i} = V_{\boldsymbol{a}, i} \odot \boldsymbol{a}$ and $\boldsymbol{e}_{r, i} = V_{\boldsymbol{e}, i} \odot \boldsymbol{e}$ denote the region-aware feature at position $\mu_i$ in the region attention mechanism, calculated by the attention vectors $V_{\boldsymbol{a}, i}$ and $V_{\boldsymbol{e}, i}$ with the Hadamard product $\odot$.

During the optimization, we apply the Incremental Sampling strategy for the lips action and eye-blinking. Specifically, we measure the lips opening degree by the height of the mouth area according to the detected facial landmarks, and use AU45 \cite{Ekman1978FacialAC} to describe the degree of eye close. Then, we gradually move the sliding window to guide the face branch to learn the deformations of the lips from close to open and eyes from open to close.

\vspace{4pt}\noindent\textbf{Inside Mouth Branch. }
The inside mouth branch represents the audio-driven dynamic inside mouth region in 3D space. Considering the inside mouth moves in a much simpler manner and is only driven by audio, we use a lightweight deformable Gaussian field to build this branch. In particular, we only predict the translation $\Delta \mu_i$ conditioned by the audio feature $\boldsymbol{a}$ for the $i$-th primitive:
\begin{equation}
    \setlength{\abovedisplayskip}{4pt}
    \setlength{\belowdisplayskip}{4pt}
    \delta_i^\text{M} = \{\Delta \mu_i^\text{M}\} = \text{MLP}(\mathcal{H}^\text{M}(\mu_i) \oplus \boldsymbol{a}).
\end{equation}

To get a better reconstruction quality of the teeth part, we apply an incremental sampling strategy that smooths the learning of the overlapping between teeth and lips with the quantitative metric AU25 \cite{Ekman1978FacialAC}.

\vspace{4pt}\noindent\textbf{Rendering. }
The final talking head image is fused with the two rendered face and inside mouth images. Based on the physical structure, we assume the rendering results from the Inside Mouth Branch are behind that from the Face Branch. Therefore, the talking head color $\mathcal{C}_\text{head}$ of pixel $\mathbf{x}_p$ can be rendered by:
\begin{equation}
    \setlength{\abovedisplayskip}{4pt}
    \setlength{\belowdisplayskip}{4pt}
\label{eq:fusion}
    \mathcal{C}_\text{head}(\mathbf{x}_p) = \mathcal{C}_\text{face}(\mathbf{x}_p) \times \mathcal{A}_\text{face}(\mathbf{x}_p) 
                    + \mathcal{C}_\text{mouth}(\mathbf{x}_p) \times (1 - \mathcal{A}_\text{face}(\mathbf{x}_p)), 
\end{equation}
where  $\mathcal{C}_\text{face}$ and $\mathcal{A}_\text{face}$ denote the predicted face color and opacity from the face branch, and $\mathcal{C}_\text{mouth}$ is the color predicted by the inside mouth branch.

\subsection{Training Details}
We keep the basic 3DGS optimization strategies to train our framework. The full process can be divided into three stages, of which the first two stages are individually applied for the two branches and the last stage is for fusion.

\vspace{4pt}\noindent\textbf{Static Initialization. }
At the beginning of the training, we first conduct an initialization via the vanilla 3DGS for the Persistent Gaussian Fields to get a coarse head structure. Following 3DGS, we use a pixel-wise L1 loss and a D-SSIM term to measure the error between the image $\hat{\mathcal{I}}_C$ rendered by parameters $\theta_C$ and the masked ground-truth image $\mathcal{I}_\text{mask}$ for each branch:
\begin{equation}
 \setlength{\abovedisplayskip}{5pt}
\setlength{\belowdisplayskip}{5pt}
    \mathcal{L}_C = \mathcal{L}_1(\hat{\mathcal{I}}_C, \mathcal{I}_\text{mask}) + \lambda \mathcal{L}_{\mathrm{D-SSIM}}(\hat{\mathcal{I}}_C, \mathcal{I}_\text{mask}) .
\end{equation}

\vspace{0pt}\noindent\textbf{Motion Learning. }
After the initialization, we add the motion fields into training via its predicted deformation $\delta$. In practice, we take the deformed parameters $\theta_D$ from Equation \ref{eq:deform_param} as the input for the 3DGS rasterizer to render the output image $\hat{\mathcal{I}}_D$. The loss function is:
\begin{equation}
\setlength{\abovedisplayskip}{5pt}
\setlength{\belowdisplayskip}{5pt}
    \mathcal{L}_D = \mathcal{L}_1(\hat{\mathcal{I}}_D, \mathcal{I}_\text{mask}) + \lambda \mathcal{L}_{\mathrm{D-SSIM}}(\hat{\mathcal{I}}_D, \mathcal{I}_\text{mask}) .
\end{equation}

\vspace{0pt}\noindent\textbf{Fine-tuning. }
Finally, a color fine-tuning stage is conducted to better fuse the head and inside mouth branches. We calculate the reconstruction loss between the fused image $\hat{\mathcal{I}}_\text{head}$ rendered by Equation \ref{eq:fusion} and the ground-truth video frame $\hat{\mathcal{I}}$ with pixel-wise L1 loss, D-SSIM, and LPIPS terms:
\begin{equation}
\setlength{\abovedisplayskip}{5pt}
\setlength{\belowdisplayskip}{5pt}
    \mathcal{L}_F = \mathcal{L}_1(\hat{\mathcal{I}}_\text{head}, \mathcal{I}) + \lambda \mathcal{L}_{\mathrm{D-SSIM}}(\hat{\mathcal{I}}_\text{head}, \mathcal{I}) + \gamma \mathcal{L}_{\mathrm{LPIPS}}(\hat{\mathcal{I}}_\text{head}, \mathcal{I}).
\end{equation}
At this stage, we only update the color parameter $f \in \theta_C$ and stop the densification strategy of 3DGS for stability.

\section{Experiment}

\subsection{Experimental Settings}\label{sec:setting}

\noindent\textbf{Dataset. } 
We collect four high-definition speaking video clips from previous publicly-released video sets~\cite{guo2021ad, li2023efficient, ye2023geneface} to build the video datasets for our experiments, including three male portraits "Macron", "Lieu", "Obama", and one female portrait "May". The video clips have an average length of about 6500 frames in 25 FPS with a center portrait, three  ("May", "Macron", and "Lieu") of which are cropped and resized to $512\times512$ and one  ("Obama") to $450\times450$. 

\vspace{4pt}\noindent\textbf{Comparison Baselines. }
In the experiments, we mainly compare our method with the most related NeRF-based methods AD-NeRF \cite{guo2021ad}, DFRF \cite{shen2022dfrf}, RAD-NeRF \cite{tang2022rad}, GeneFace \cite{ye2023geneface} and ER-NeRF \cite{li2023efficient}, which render talking head via person-specific radiance fields trained with speech videos. Additionally, we also take the state-of-the-art 2D generative models (Wav2Lip \cite{prajwal2020wav2lip}, IP-LAP \cite{zhong2023identity} and DINet \cite{zhang2023dinet}), which do not need person-specific training, and person-specific methods (SynObama \cite{suwajanakorn2017synthesizing}, NVP \cite{thies2020nvp}, and LSP \cite{lu2021lsp}) as the baselines.

\vspace{4pt}\noindent\textbf{Implementation Details. }
Our method is implemented on PyTorch. For a specific portrait, we first train both the face and inside mouth branches for $50,000$ iterations parallelly and then jointly fine-tune them for $10,000$ iterations. Adam \cite{kingma2014adam} and AdamW \cite{loshchilov2018adamw} optimizers are used in training. In the loss functions, $\lambda$ and $\gamma$ are set to $0.2$ and $0.5$. All experiments are performed on RTX 3080 Ti GPUs. The overall training process takes about 0.5 hours. A pre-trained DeepSpeech model \cite{hannun2014deepspeech} is used as a basic audio feature extractor.

\subsection{Quantitative Evaluation}

\noindent\textbf{Comparison Settings. }
To evaluate the reconstruction quality and lip-audio synchronization ability, our quantitative comparison contains two settings: \textbf{1)} The \emph{self-reconstruction setting}, where we split each of the four videos into training and test sets, and use the audio, expression, and pose sequences in the unseen test set to reconstruct the talking head in a self-driven way for quality evaluation.  \textbf{2)} The \emph{lip-synchronization setting}, where we use the audio track from other videos to drive the models trained in the first setting and evaluate lip-synchronization, focusing on situations with cross-domain input audios. Specifically, we use the same audio samples as previous works \cite{li2023efficient, peng2023synctalk} from NVP and SynObama as two test audios A and B to evaluate the "Obama" and "May" portraits. Since both audio A and B are from unseen videos with male voices, the evaluation results, especially on "May" with a different gender, can well illustrate the generalization ability. Tests for NVP, SynObama, and SSP-NeRF are conducted only on their released demos due to the lack of training codes.

\vspace{4pt}\noindent\textbf{Metrics and Measurements.}
In the aspect of static image quality, we employ PSNR for the overall quality, LPIPS \cite{zhang2018lpips} for high-frequency details, and SSIM \cite{wang2004ssim} to evaluate face structure. For dynamic motions, we also utilize the landmark distance  (LMD) \cite{chen2018lmd}  and the confidence score (Sync-C) and error distance (Sync-D) of SyncNet \cite{chung2017syncnet1, chung2017syncnet2} for lip synchronization. Additionally, we estimate the action units \cite{Ekman1978FacialAC} of the videos by OpenFace \cite{baltrusaitis2018openface, baltruvsaitis2015openface2} and divide them into an upper-face action unit error (AUE-U) and lower-face action unit error (AUE-L) according to their definitions to separately evaluate the upper-face and mouth motions.

In the first setting, we measure the PSNR and LPIPS on the whole image, and SSIM on the face region. We also record the person-specific training time and inference FPS of all methods. Since all the 2D-based generative baselines do not modify the upper part of the face, we do not measure their AUE-U. Notably, we provide another video clip as the image reference for Wav2Lip to avoid information leakage, for which PSNR, LPIPS, and SSIM are not valid. In the second setting, we use the non-comparison-based Sync-C and Sync-E to quantitatively measure the lip-synchronization quality.

\begin{table}[t]
    \caption{\textbf{The quantitative results of the \emph{self-reconstruction setting}}. The best and second-best methods are in \textbf{bold} and {\ul underline}, respectively.}
    \label{tab:setting1}
    \vspace{-2mm}
\resizebox{1\linewidth}{!}{
        \setlength{\tabcolsep}{3.7mm}
        \centering
        \begin{tabular}{lccc|ccc|cc}
        \toprule
        \multirow{2}{*}{Methods} & \multicolumn{3}{c|}{Rendering Quality} & \multicolumn{3}{c|}{Motion Quality} & \multicolumn{2}{c}{Efficiency} \\ 

                & PSNR $\uparrow$ & LPIPS $\downarrow$ & SSIM $\uparrow$ & LMD $\downarrow$ & AUE-(L/U) $\downarrow$ & Sync-C $\uparrow$ & Time & FPS \\ 
        Ground Truth  & N/A            & 0               & 1.000              & 0              & 0/0              & 7.584          & -   & -   \\ \midrule
        Wav2Lip \cite{prajwal2020wav2lip}      
                & -               & -               & -             & 6.861           & 1.46 / -      & \textbf{8.749} & -   & 21.6 \\

        IP-LAP \cite{zhong2023identity}      
                & \textbf{35.34} & 0.0405          & {\ul 0.903}        & 5.601           & 0.77 / -     & 4.897         & -        & 3.18 \\

        DINet \cite{zhang2023dinet}      
                & 32.08          & 0.0393          & 0.856             & 6.411            & 0.97 / -     & 6.321   & -     & 27.2 \\ \midrule

        AD-NeRF \cite{guo2021ad}      
                & 31.87          & 0.0942          & 0.877              & 2.791          & 0.71/1.26              & 5.353      & 18.7h  & 0.11 \\
        DFRF \cite{shen2022dfrf}      
                & 31.73          & 0.0858          & 0.876              & 3.406          & 0.74/1.40              & 4.127      & 22.4h  & 0.04 \\

        RAD-NeRF \cite{tang2022rad}     
                & 33.07          & 0.0530          & 0.887              & 2.761          & 0.65/1.14              & 5.052      & 5.3h  & 28.7     \\ 
                
        GeneFace \cite{ye2023geneface}   
                & 30.49          & 0.0670          & 0.846              & 3.339          & 1.28/1.34              & 5.291      & 5.8h   & 20.9  \\

        ER-NeRF \cite{li2023efficient}
                & 32.83          & 0.0289          & 0.889              & 2.676          & {\ul 0.55}/0.88      & 5.295      & {\ul 2.1h}  & {\ul 31.2} \\ 
                
        ER-NeRF$+\boldsymbol{e}$
                & 33.14          & {\ul 0.0271}          & 0.902              & {\ul 2.623}          & 0.57/{\ul 0.31}      & 5.754      & -           & -          \\ \midrule 
        \textbf{Ours}
                & {\ul 33.61}    & \textbf{0.0259} & \textbf{0.910}              & \textbf{2.586} & \textbf{0.53}/\textbf{0.22}
              & {\ul 6.516}      & \textbf{0.5h}  & \textbf{108} \\ \bottomrule 
        \end{tabular}
}

\end{table}

\begin{table}[t]
\caption{\textbf{The quantitative results of the \emph{lip-synchronization setting}}. The best and second-best methods are in \textbf{bold} and {\ul underline}, respectively.}
\label{tab:setting2}
\vspace{-2mm}
\resizebox{1\linewidth}{!}{
\centering
\setlength{\tabcolsep}{3.5mm} 
\begin{tabular}{lcccc|cccc}
\toprule
\multirow{3}{*}{Method}   & \multicolumn{4}{c}{Test Audio A} & \multicolumn{4}{c}{Test Audio B} \\
             & \multicolumn{2}{c}{"Obama"} & \multicolumn{2}{c}{"May"} & \multicolumn{2}{c}{"Obama"} & \multicolumn{2}{c}{"May"} \\   \cmidrule(l){2-9} 

             & Sync-E $\downarrow$   & Sync-C $\uparrow$ & Sync-E $\downarrow$   & Sync-C $\uparrow$  & Sync-E $\downarrow$   & Sync-C $\uparrow$  & Sync-E $\downarrow$   & Sync-C $\uparrow$  \\
Ground Truth & 0        & 6.701       & 0       & 6.701        & 0       & 7.309      & 0       & 7.309 \\ \midrule

LSP \cite{lu2021lsp}          
                                        & {\ul 8.683}        & 5.045        & {\ul 9.511}         & 4.441               
                                        & {\ul 8.640}        & 5.504        & 9.882         & 4.167          \\ 
SynObama \cite{suwajanakorn2017synthesizing}
                                        & 8.197        & \textbf{6.802}     & -             & -     
                                        & -            & -                  & -             & -      \\
NVP \cite{thies2020nvp}          
                                        & -            & -                  & -             & -      
                                        & 10.175       & 4.316              & -             & -       \\ \midrule
AD-NeRF \cite{guo2021ad}     
                                        & 9.742        & 5.195           & 9.517        & {\ul 4.757}       
                                        & 10.682       & 4.314          & {\ul 9.518}   & {\ul 5.319}             \\
DFRF \cite{shen2022dfrf}         
                                        & 10.662        & 3.905                  & 10.830       & 3.135                                  
                                        & 11.044        & 3.690                  & 11.248       & 3.215                             \\
RAD-NeRF \cite{tang2022rad}     
                                        & 9.552         & 5.585             & 11.883             & 2.000           
                                        & 8.680         & 6.667             & 11.176             & 2.426         \\ 

GeneFace \cite{ye2023geneface}          & 9.052       & 5.336          & 10.259         & 3.569                   
                                        & 8.966       & 5.674          & 10.173         & 4.280                 \\

ER-NeRF \cite{li2023efficient}     
                                        & 9.123         & {\ul 6.134}          & 10.251       & 3.639             
                                        & 8.688         & {\ul 6.706}    & 10.535       & 4.141            \\ 
                                        
ER-NeRF$+\boldsymbol{e}$     
                                        & 9.573         & 6.092          & 9.825       & 4.012             
                                        & 8.934         & 6.577          & 11.226       & 4.423            \\ \midrule
\textbf{Ours}     
                                        & \textbf{8.635}  & 5.962     & \textbf{9.368}    & \textbf{4.774}     
                                        & \textbf{8.627}         & \textbf{6.737}          & \textbf{9.273}     & \textbf{5.441}            \\ \bottomrule
\end{tabular}
}
\vspace{-2mm}
\end{table}

\vspace{4pt}\noindent\textbf{Evaluation Results. }
We report the results of the two settings in Table \ref{tab:setting1} and Table \ref{tab:setting2}, respectively. Considering the upper-face expression condition would also influence performance \cite{li2023efficient, peng2023synctalk}, we add our upper-face feature $\boldsymbol{e}$ to ER-NeRF \cite{li2023efficient} as "ER-NeRF+$\boldsymbol{e}$" for a fair comparison. 
\textbf{1)} In the \emph{self-reconstruction setting}, our method achieves the best overall image quality, motion quality, and efficiency. For image quality, our method performs best in rendering accurate details (LPIPS) and structure (SSIM), due to our deformation-based motion representation and persistent head structure. 
In the aspect of motion quality, our method outperforms all NeRF methods in all metrics. Notably, TalkingGaussian gets a Sync-C score even higher than the generative method IP-LAP and DINet, demonstrating the powerful modeling ability of our method. Although Wav2Lip gets the best scores in Sync-C, its shortcomings in preserving personal talking styles lead to poor AUE-L and LMD. Moreover, due to the efficiency improvement brought by 3DGS, our method reaches the fastest training and inference speed in all baselines.
\textbf{2)} In the results in the \emph{lip-synchronization setting}, our method shows the best generalization performance. Although ER-NeRF can also get good scores on "Obama", it performs much worse in the more challenging cross-gender situation of "May". This phenomenon can also be observed in many other NeRF-based methods, especially RAD-NeRF which keeps a complex audio encoding module. Surprisingly, AD-NeRF performs well in this situation, despite having a relatively blurry rendering. Besides the difference in model quality, we consider this decreasing generalizability also to be caused by the overfitting of the audio feature when previous NeRF-based methods try to fit the unsmooth changing appearance to reconstruct delicate talking heads. Instead, when just using the same unimproved audio feature extractor as most previous baselines \cite{thies2020nvp, guo2021ad, shen2022dfrf, li2023efficient}, our method can simultaneously keep high-level static rendering quality and best generalization ability for various training videos and input audios, thanks to our simpler and smoother deformation-based motion representation.

\subsection{Qualitative Evaluation}

\begin{figure}[t]
    \centering
    \includegraphics[width=\linewidth]{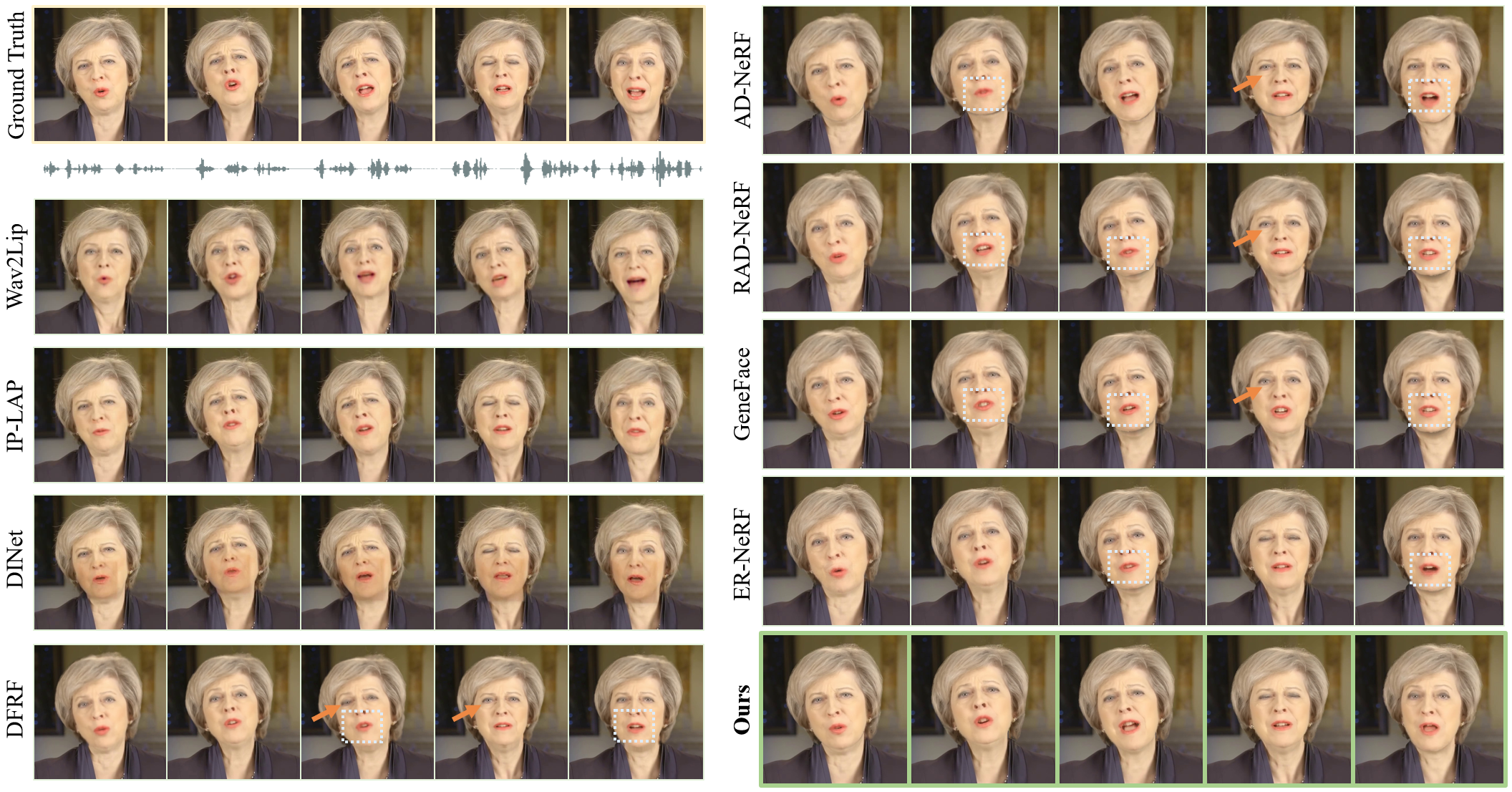}
    \vspace{-3mm}
    \caption{\textbf{Qualitative comparison of visual-audio synchronization.} Our method performs best in synthesizing accurately synchronized talking head compared with all baselines \cite{prajwal2020wav2lip, zhong2023identity, zhang2023dinet, guo2021ad, shen2022dfrf, tang2022rad, ye2023geneface, li2023efficient}. Please \textbf{zoom in for better visualization.}}
    \label{fig:quali1}
    \vspace{-3mm}
\end{figure}

\begin{figure}[t]
    \centering
    \includegraphics[width=\linewidth]{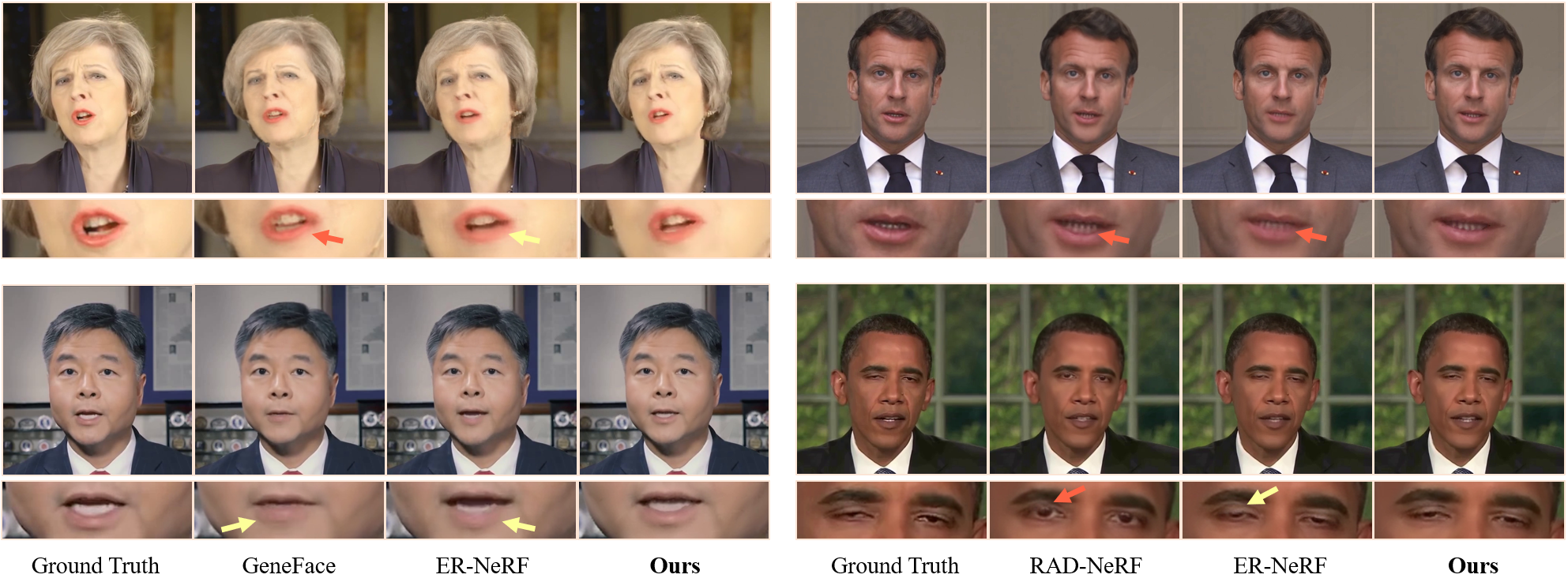}
    \vspace{-5mm}
    \caption{\textbf{Qualitative comparison of the generated facial details.} Our method synthesizes more accurate and intact details than the recent NeRF-based state-of-the-art methods \cite{ye2023geneface, tang2022rad, li2023efficient}. Please \textbf{zoom in for better visualization.}}
    \label{fig:quali2}
    \vspace{-1mm}
\end{figure}

\noindent\textbf{Evaluation Results. } 
To qualitatively evaluate the synthesis quality, we show the keyframes of a reconstructed sequence from the \emph{self-reconstruction setting} and details of four portraits in Fig.~\ref{fig:quali1} and \ref{fig:quali2}. 
\textbf{1)} Comparing the synthesized motion sequence in Fig.~\ref{fig:quali1}, our TalkingGaussian outperforms other methods by generating better visual-audio synchronized results. The generative methods (Wav2Lip, IP-LAP, and DINet) are short in generating high-quality images, as the trade-off for their one or few-shot ability. Lack of precise control signals, most NeRF-based baselines can not control audio-independent actions like eye blinking (orange arrow).  
While most NeRF-based baselines fail to synthesize some difficult mouth actions (blue box), our TalkingGaussian can precisely reappear these motions, without introducing any advanced audio encoders \cite{ye2023geneface, peng2023synctalk}. This demonstrates the effectiveness of the learning task simplifications brought by our two decomposition designs.
\textbf{2)} The comparison of synthesized details in Fig. \ref{fig:quali2} shows our advantages in facial fidelity and fidelity. As illustrated in Sec. \ref{sec:fields}, both RAD-NeRF and ER-NeRF exhibit distorted (red arrow) and blurry (yellow arrow) facial features in the dynamic regions. Even using the more precise facial landmarks to condition the NeRF renderer, GeneFace still can't escape from this inherent trouble brought by the previous appearance-modification paradigm. By purely representing motions with deformation, our method tackles this problem and succeeds in synthesizing more accurate and intact facial features.

\vspace{4pt}\noindent\textbf{User Study. } 
To better judge the visual quality in real scenarios judged by humans, we conducted a user study where a total of 32 talking head videos were generated by 8 methods. Then we invited 16 attendees to rate these methods according to their generated results from three aspects: (1) Lip-sync Accuracy; (2) Video Realness; and (3) Image Quality. The results are reported in Table \ref{tab:user_study}, in which our TalkingGaussian performs the best in all three aspects, demonstrating the potential value of our method in real-world applications.

\begin{table}[t]

\caption{\textbf{User Study.} The rating is in the range of 1-5, higher denotes better. We highlight the \textbf{best} and {\ul second best} results.}
\label{tab:user_study}
\vspace{-2mm}
\resizebox{1\linewidth}{!}{
\centering
\begin{tabular}{lcccccccc}
\toprule
Methods           & Wav2Lip \cite{prajwal2020wav2lip} & IP-LAP \cite{zhong2023identity} & DI-Net \cite{zhang2023dinet} & AD-NeRF \cite{guo2021ad}  & GeneFace \cite{ye2023geneface} & RAD-NeRF \cite{tang2022rad} & ER-NeRF \cite{li2023efficient}  & \textbf{TalkingGaussian}  \\ \midrule
Lip-sync Accuracy & 2.50     & 1.63     & 3.25      & 2.75      & 3.13      & 3.19      & {\ul 3.56}      & \textbf{3.94}     \\
Image Quality     & 1.75     & 2.44     & 2.69      & 3.25      & {\ul 3.69}      & 3.31      & 3.63      & \textbf{4.06}        \\
Video Realness    & 1.69     & 1.88     & 1.88      & 3.19      & 3.31      & 3.19      & {\ul 3.44}      & \textbf{3.88}     \\  \bottomrule
\end{tabular}
}
\vspace{-2mm}
\end{table}

\subsection{Ablation Study}

\begin{figure}[t]
    \centering
    \includegraphics[width=0.98\linewidth]{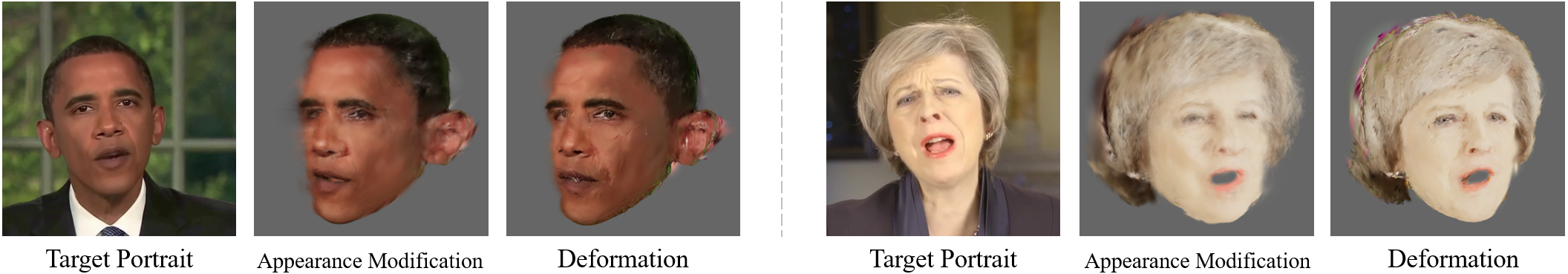}
    \vspace{-2mm}
    \caption{\textbf{3D visualization} of the heads generated by deformation and appearance modification on 3DGS. Deformation performs better in generating more precise geometry. }
    \label{fig:3dhead}
\end{figure}

\begin{table}[t]
    \caption{\textbf{Ablation Study} of our contributions under the self-reconstruction setting.}
    \label{tab:ablation}
    \vspace{-2mm}
    \centering
    \resizebox{1\linewidth}{!}{    
    \setlength{\tabcolsep}{2mm} 
\begin{tabular}{@{}c|cc|cc|ccc|ccc@{}}
\toprule
Backbone                  & Appearance           & Deformation & FMD        & IS                    & PSNR $\uparrow$ & LPIPS $\downarrow$ & SSIM $\uparrow$ & LMD $\downarrow$ & AUE-(L/U) $\downarrow$ & Sync-C $\uparrow$ \\ \midrule
\multirow{2}{*}{Tri-Hash} & \checkmark           &             &            &                       & 33.14           & 0.0271             & 0.902           & 2.623            & 0.57/0.31             & 5.754             \\
                          &                      & \checkmark  &            & \checkmark            & 31.50           & 0.0334             & 0.877           & 3.016            & 0.67/0.38             & 5.285             \\ \midrule
\multirow{5}{*}{3DGS}     & \checkmark           &             &            &                       & 33.34           & 0.0355             & 0.904           & 2.630            & 0.56/0.25             & 6.001             \\
                          &                      & \checkmark  &            & \checkmark            & 33.42           & 0.0290             & 0.903           & 2.665            & 0.54/0.23             & 5.676             \\ \cmidrule(l){2-11} 
                          & \checkmark           &             & \checkmark &                       & 33.27           & 0.0351             & 0.904           & 2.605            & 0.55/0.24             & 6.332             \\
                          & \multicolumn{1}{l}{} & \checkmark  & \checkmark & \multicolumn{1}{l|}{} & {\ul 33.57}           &{\ul  0.0260}             & {\ul 0.906}           & \textbf{2.584}   & \textbf{0.53}/{\ul 0.23}    & {\ul 6.497}       \\ 
                          &                      & \checkmark  & \checkmark & \checkmark            & \textbf{33.61}  & \textbf{0.0259}    & \textbf{0.910}  & {\ul 2.586}      & \textbf{0.53/0.22}    & \textbf{6.516}    \\ \bottomrule
\end{tabular}
    }
\vspace{-3mm}
\end{table}

To prove the effectiveness of our contributions, we conduct the ablation study using the self-reconstruction setting. The results are reported in Table \ref{tab:ablation}.

\vspace{4pt}\noindent\textbf{Motion Representation. } 
First, we use the same module as our motion fields to predict the deformation and appearance modification for the Tri-Hash backbone from ER-NeRF \cite{li2023efficient} and 3DGS \cite{kerbl2023gaussian} to evaluate the motion representations. Due to the lack of accurate point-wise controls, deformation performs worse on the NeRF-based Tri-Hash. After introducing 3DGS, deformation shows its advantage in preserving better facial features and bringing higher image quality scores. We also visualize the results on 3DGS with the same conditions in Fig. \ref{fig:3dhead} for comparison. The results demonstrate the effectiveness of both our 3DGS-based persistent head structure and deformation-based motion representation. 

\vspace{4pt}\noindent\textbf{Face-Mouth Decomposition} (FMD).  We apply our FMD to the 3DGS backbone to illustrate its effect. Although the results show that FMD can help in lip-synchronization in all situations, the improvement is much larger when combined with deformation, since it can solve the face-mouth motion conflict that seriously affects deformation learning. With this combination, our framework successfully reaches the best motion quality, especially in lip-synchronization.

\vspace{4pt}\noindent\textbf{Incremental Sampling} (IS). Raised by the unstable optimization process of deformation,  some jitters, incomplete motions, and errors in geometry structure can be observed in the generated videos, which lead to a lower SSIM. These problems are relieved after applying IS, demonstrating its contribution to generating more smooth and realistic talking heads. 

\vspace{-1mm}
\section{Conclusion} 
\vspace{-1mm}

This paper represents a novel deformation-based framework TalkingGaussian for high-quality 3D talking head synthesis. Our study is the first to reveal a "facial distortion" problem caused by inaccurate predictions of the rapidly changing appearance. By maintaining a persistent head structure with 3DGS and decomposing inconsistent motions into different spaces, TalkingGaussian addresses this problem with a deformation paradigm and achieves superior performance in synthesizing realistic and accurate talking heads compared to existing methods.

\vspace{2pt}\noindent\textbf{Ethical Consideration. } We hope our method can promote the healthy development of digital industries. However, it must be noted that our method may be misused for malicious purposes and cause negative influence. We recommend the responsible use of this technique. As part of our responsibility, we will also assist in developing deepfake detection techniques by sharing our generated results.

\newpage

\title{Supplementary Material for TalkingGaussian} 
\author{Jiahe Li\inst{1} \and
Jiawei Zhang\inst{1} \and
Xiao Bai\inst{1}\thanks{Corresponding author: Xiao Bai (baixiao@buaa.edu.cn).} \and
Jin Zheng\inst{1} \and
Xin Ning\inst{2} \and
Jun Zhou\inst{3} \and
Lin Gu\inst{4,5}
}

\authorrunning{J. Li et al.}

\institute{School of Computer Science and Engineering, State Key Laboratory of \\ Complex \& Critical Software Environment, Jiangxi Research Institute, \\ Beihang University \and
Institute of Semiconductors, Chinese Academy of Sciences \and
School of Information and Communication Technology, Griffith University \and
RIKEN AIP \and 
The University of Tokyo}

\maketitle
\renewcommand\thesection{\Alph{section}}
\setcounter{figure}{7}
\setcounter{table}{4}

\section*{Overview}
In the supplementary material, we first report additional experiments in Sec. \ref{sec:exp} and implementation details in Sec. \ref{sec:impl}. We further show an additional visualization in Sec. \ref{sec:vis}, and discuss the responsibility to human subjects and ethical consideration in Sec. \ref{sec:data} and \ref{sec:ethi}. Limitations and future work are summarized in Sec. \ref{sec:discuss}. A supplementary video is also provided as an additional illustration.

\section{Additional Experiments \label{sec:exp}}
\subsection{Hybrid Motion Representation} 
We additionally conduct an experiment to explore whether a hybrid representation of mixing deformation and appearance modification could benefit the performance. The results are reported in Table \ref{tab:ab_supp}. Upon the basic deformation $\delta$, we first evaluate the setting of additionally predicting a factor to adjust the opacity $\alpha$. Then we further try to predict the RGB color of each primitive directly rather than using the SH feature $f$. The results show that: 1) Adding $alpha$ would not help more. This is reasonable since the opacity of each primitive should be a static value. 2) Adding RGB would result in a blurry rendering. In the first aspect, it would bring a larger burden for the network to store more information. On the other hand, a non-persistent color can still suffer from the distortion problem from inaccurate appearance prediction to some extent. 

\begin{table}[t]
    \caption{Exploration of hybrid motion representations.}
    \label{tab:ab_supp}
    \centering
    \resizebox{0.6\linewidth}{!}{
        \setlength{\tabcolsep}{5mm}
        \centering
        \begin{tabular}{@{}lccc@{}}
        \toprule
        Settings & PSNR $\uparrow$ & LPIPS $\downarrow$ & SSIM $\uparrow$ \\ \midrule

        $\delta  + \alpha$
                & 33.60    & 0.0261 & 0.908       \\ 
        $\delta  + \alpha + \text{RGB}$
                & \textbf{33.63}    & 0.0264 & 0.907       \\ \midrule
        $\delta$ 
                & 33.61    & \textbf{0.0259} & \textbf{0.910}      \\  \bottomrule 
        \end{tabular}
}
\end{table}

\subsection{Extension} 
Besides the experiment settings in the main paper, our method is scalable and can extend to a wider range of applications. 

\vspace{6pt}\noindent\textbf{Audio Feature Extractor.} 
Following previous baselines, we have adopted a pre-trained DeepSpeech \cite{hannun2014deepspeech} model to extract audio features in the main experiments, for a fair comparison. In fact, our method can also easily connect to more powerful feature extractors and become stronger. Table \ref{tab:encoder} shows our performance with different audio feature extractors Wav2Vec 2.0 \cite{baevski2020wav2vec} and HuBERT \cite{hsu2021hubert} under the \textit{self-reconstruction setting}. The results demonstrate a high-quality audio feature could boost the performance of TalkingGaussian, especially on lip-synchronization, showing the growth potential of our approach.

\begin{table}[t]
    \caption{Exploration of adopting different audio encoders under \textit{self-reconstruction setting}. The best and second-best results are in \textbf{bold} and {\ul underline}.}
    \label{tab:encoder}
    \resizebox{1\linewidth}{!}{
        \setlength{\tabcolsep}{3.7mm}                                       
        \centering
        \begin{tabular}{@{}lccc|ccc@{}}
        \toprule
        \multirow{2}{*}{Extractor} & \multicolumn{3}{c|}{Rendering Quality} & \multicolumn{3}{c}{Motion Quality} \\ 

                & PSNR $\uparrow$ & LPIPS $\downarrow$ & SSIM $\uparrow$ & LMD $\downarrow$ & AUE-(L/U) $\downarrow$ & Sync-C $\uparrow$\\ 
        Ground Truth  & N/A            & 0               & 1.000              & 0              & 0/0              & 7.584      \\ \midrule
        DeepSpeech
                & \textbf{33.61}    & {\ul 0.0259} & {\ul 0.910}     & 2.586     & 0.53/\textbf{0.22}  & 6.516     \\  
        Wav2Vec 2.0
                & 33.59    & 0.0260 & \textbf{0.911}     & \textbf{2.582}     & \textbf{0.52}/{\ul 0.23}  & {\ul 6.552}    \\ 
        HuBERT
                & {\ul 33.60}    & \textbf{0.0258} & 0.909     & {\ul 2.583}     & \textbf{0.52}/0.24  & \textbf{6.667}   \\ \bottomrule 
        \end{tabular}
}
\end{table}

\vspace{6pt}\noindent\textbf{Cross-Lingual and Cross-Gender.} 
Our method can also applied to more challenging cross-lingual and cross-gender cases. In this experiment, we collect 4 training videos, in which 2 males and 2 females with English and French audio are included, and use challenging 3 test audio clips to drive their corresponding models. The three test audios consist of two German audio clips separately with a female and a male voice, and a Chinese audio clip with a male voice. In this setting, we compare our method to two baselines ER-NeRF \cite{li2023efficient} and GeneFace \cite{ye2023geneface} that utilize different extractors. 

In Figure \ref{tab:crossling}, the results show that our models all perform better than the baselines while using the same extractors. Notably, GeneFace has further used the HuBERT features to pre-train an intermediate representation on a large audio-visual corpus to enhance its generalizability for cross-domain audios. The generated results have been provided in the supplementary video.

\begin{table}[!t]
    \caption{Exploration of cross-lingual and cross-gender situations. The best results for each audio feature extractor are in \textbf{bold}. }
    \label{tab:crossling}
    \resizebox{1\linewidth}{!}{
        \setlength{\tabcolsep}{2.5mm}
        \centering
        \begin{tabular}{@{}llcc|cc|cc@{}}
        \toprule
        \multirow{2}{*}{Extractor} & \multirow{2}{*}{Methods} & \multicolumn{2}{c|}{\textit{Female, German}} & \multicolumn{2}{c|}{\textit{Male, German}} & \multicolumn{2}{c}{\textit{Male, Chinese}}\\

                    &             & Sync-E $\downarrow$ & Sync-C $\uparrow$ & Sync-E $\downarrow$ & Sync-C $\uparrow$ & Sync-E $\downarrow$ & Sync-C $\uparrow$ \\  \midrule 
        \multirow{2}{*}{DeepSpeech} & ER-NeRF
                & 9.773    & 3.273 & 10.497     & 3.381     & 10.577  & 2.893     \\  
                                    & \textbf{Ours}
                & \textbf{9.399}    & \textbf{3.720} & \textbf{9.677}     & \textbf{4.407}     & \textbf{10.322}  & \textbf{3.378}   \\ \midrule
        \multirow{2}{*}{HuBERT} & GeneFace
                & 8.753    & 4.059 & 9.208     & 4.969     & 10.597  & 3.720    \\ 
                                    & \textbf{Ours}
                & \textbf{8.260}    & \textbf{4.691} & \textbf{8.323}     & \textbf{5.556}     & \textbf{8.856}  & \textbf{4.539}   \\ \bottomrule 
        \end{tabular}
}
\end{table}

\vspace{6pt}\noindent\textbf{Singing.} 
While inputting a song, our method can even synthesize high-quality singing talking heads with no such training audio included. This demonstrates our surprising generalization ability and robustness for cross-domain inputs, and shows applicability for a wider range of situations. The generated videos can be found in the supplementary video.

\section{Implementation Details \label{sec:impl}}

\subsection{Model Details.} 

\vspace{0pt}\noindent\textbf{Preprocessing.} 
In the main paper, we use a frozen DeepSpeech \cite{hannun2014deepspeech} model to extract raw audio features. Then a CNN-based attention module in previous NeRF-based works \cite{guo2021ad, shen2022dfrf, tang2022rad, li2023efficient} is adopted to adapt and smooth the audio features. The upper-face expression feature is composed of a set of action units that only relate to the upper-face motions, specifically: 1, 2, 4, 5, 6, 7, and 45. We use the action units detected by OpenFace \cite{baltrusaitis2018openface, baltruvsaitis2015openface2} as the input signals. Utilizing the same preprocessor in previous works \cite{tang2022rad, li2023efficient}, we first estimate the head pose via a BFM \cite{paysan2009bfm} facial model, and then inversely calculate the camera pose.

\vspace{6pt}\noindent\textbf{Persistent Gaussian Fields.} 
The Persistent Gaussian Fields are built on the 3DGS repository \footnote{https://github.com/graphdeco-inria/gaussian-splatting}. We init the fields with a random point cloud, and inherit the adaptive density control from 3DGS during training. For rendering, we adopt a modified 3DGS rasterizer \footnote{https://github.com/ashawkey/diff-gaussian-rasterization}, which enables alpha supervision, to penalize the empty areas.

\vspace{6pt}\noindent\textbf{Grid-based Motion Fields.}
In the experiments, we use three 2D hash-encoders \cite{muller2022instant} with a 3-layer MLP decoder to implement the Grid-based Motion Fields. For each 2D hash-encoder, we set the resolution range from 16 to 256 in the face branch, and 64 to 384 for the inside mouth branch. All these encoders are set with 12 levels. The hidden dimension of the MLP decoder is set to 64 and 32, respectively for the face and inside mouth branches.

\vspace{6pt}\noindent\textbf{Optimizer. } 
During training, we keep two optimizers to optimize the Persistent Gaussian Fields and Grid-based Motion Fields separately. For the Persistent Gaussian Fields, we inherit the Adam optimizer from 3DGS with a similar hyperparameter setting to optimize the Gaussians. For the motion fields, we adopt an AdamW optimizer with learning rates of $5e-3$ and $5e-4$ for the hash encoder and other parts. An exponential scheduler is used to adjust the learning rates.

\subsection{Face-Mouth Segmentation}
\begin{figure}[t]
    \centering
    \includegraphics[width=\linewidth]{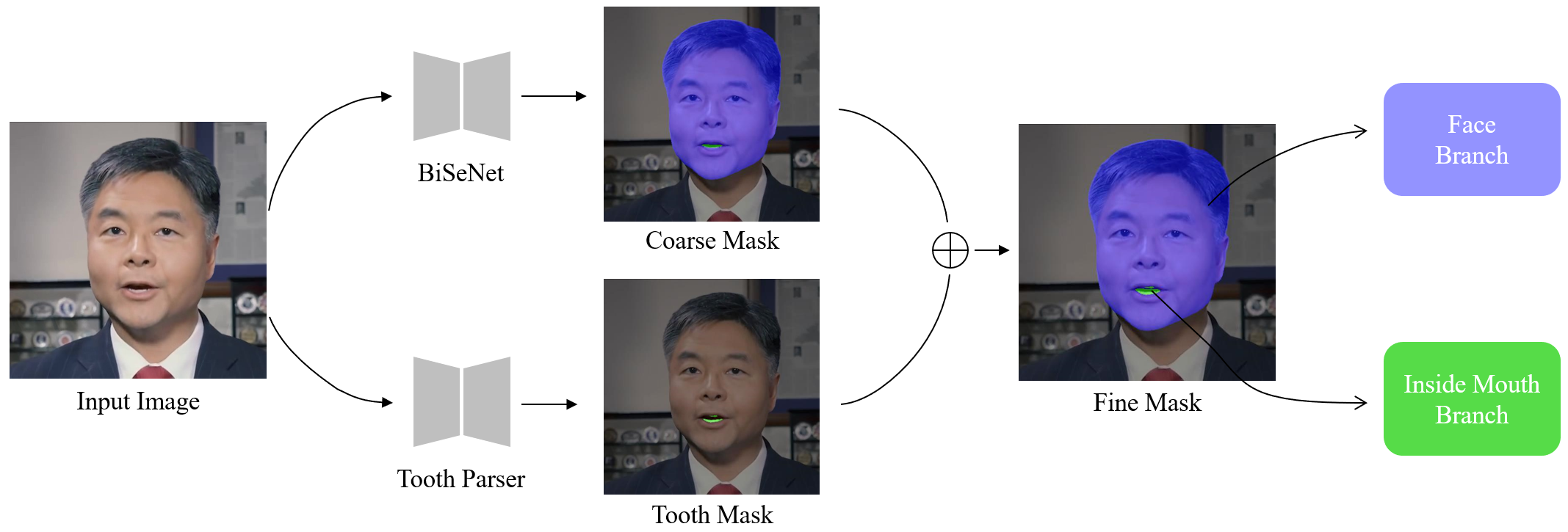}
    \caption{Illustration of face and inside mouth segmentation.}
    \label{fig:seg}
    \vspace{-3mm}
\end{figure}

As mentioned in Sec. 3.3, a semantic mask is used to divide the face and inside mouth in the 2D image. Specifically, we use a combination of two off-the-shelf face parsing models to generate the mask. First, we use a BiSeNet parser \cite{yu2018bisenet} pre-trained on CelebAMask-HQ dataset \cite{CelebAMask-HQ} to predict a coarse mask that contains a coarse mouth segmentation for the whole head and inside mouth. Due to the domain gap, the mouth mask from this face parser may fail to completely cover the mouth sometimes. To enhance the robustness, we further introduce a ResNet-based FPN trained on EasyPortrait \cite{EasyPortrait} as a tooth parser to predict a tooth mask. Finally, we overlay these two masks to get a finer one, then use the corresponding masked image as $\mathcal{I}_\text{mask}$ in Eq. (10, 11) to supervise the two branches of TalkingGaussian. For each branch, we apply the semantic mask on both the ground truth image and the predicted result during training. The overall pipeline is shown in Fig \ref{fig:seg}.

\section{Additional Visualization \label{sec:vis}}

\begin{figure}[!t]
    \centering
    \includegraphics[width=\linewidth]{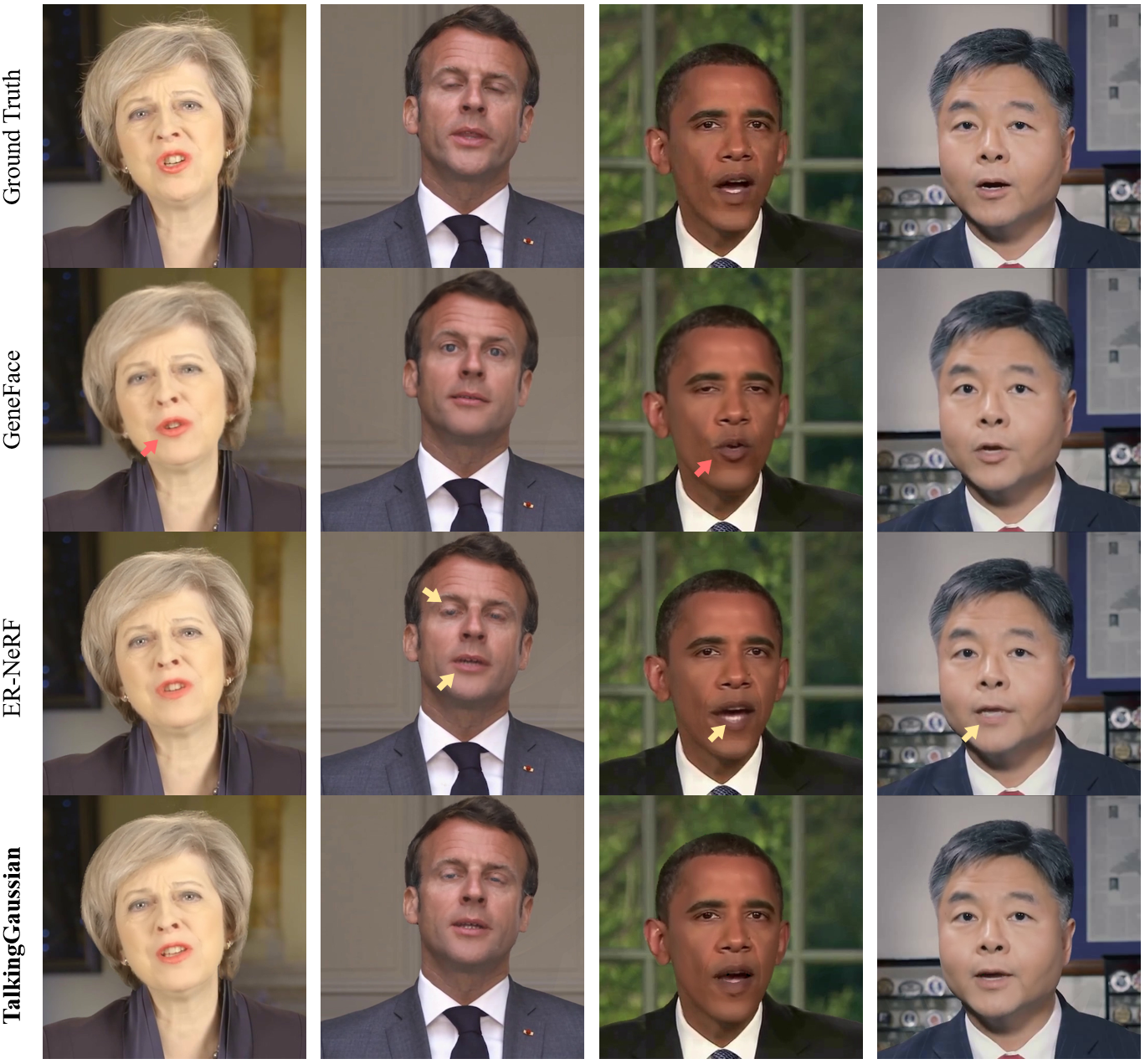}
    \caption{\textbf{Additional High-definition Comparisons.} ER-NeRF \cite{li2023efficient} heavily sufferers the facial distortion problem caused by inaccurate appearance prediction. GeneFace \cite{ye2023geneface} performs better in preserving fidelity, since it has introduced an intermediate representation to bridge the audio-visual mapping. However, its synchronization quality drops. In comparison, our method synthesizes better talking heads both in static and dynamic.}
    \label{fig:hd}
\end{figure}

Here we show some high-definition synthesized frames in Fig. \ref{fig:hd} for a convenient and intuitive visualization. Compared with current SOTA methods GeneFace \cite{ye2023geneface} and ER-NeRF \cite{li2023efficient}, our method performs best in image quality while retaining a high lip-sync accuracy. We have also provided more additional results in our \textcolor{blue}{supplementary video}. We strongly recommend watching it for better visualization.

\section{Dataset Declaration \label{sec:data}}
In the experiments, all of the multimedia datasets we used were obtained from existing works \cite{guo2021ad, li2023efficient, ye2023geneface}. To our knowledge, most of these data are collected from the internet. In our work, we have tried our best to use data containing only public figures to avoid invading personal privacy. All the data are manually checked to reduce the existence of offensive content.

\section{Ethical Consideration \label{sec:ethi}}
As our target, we hope our TalkingGaussian can promote the healthy development of digital industries. However, it must be noted that our method may be misused for malicious purposes and cause negative influence. We recommend the responsible use of this technique:

\begin{itemize}
    \item \textbf{Informed Consent.} Whenever this technique is in use for spread purposes, ensure that all individuals in the training data have provided explicit, informed consent. 
    \item \textbf{Disclosure.} Please disclose the use of our method, and any other deepfake techniques as well, in all synthesized products. This is critical to ensure all audiences are aware that the content is real, and may include misleading information.
\end{itemize}

For protection purposes, we will support the development of more powerful deepfake detectors to alert people to the presence of fake content.

\section{Limitations and Future Work \label{sec:discuss}}
In this paper, our proposed TalkingGaussian outperforms in rendering high-quality lip-synchronized talking head videos, with better facial fidelity and higher efficiency than previous methods. Despite that, our method still has some limitations. 

In the first aspect, some noisy primitives may randomly occur due to the densification operation of 3DGS. Although this can be relieved by a smoother optimization process provided by Incremental Sampling, it would still sometimes influence the quality. In future works, we will consider adding more constraints to better control the primitive's growth. 

On the other hand, the face and inside mouth branches are aligned via the audio feature in our method, enabling free and individual learning for their own motions. Nevertheless, this connection is not tight enough. Although it is sufficient for most in-domain audio inputs like speeches from the same person as that in the training data, the face and inside mouth area may be misaligned in some cross-domain situations. To solve this problem, we may build a better awareness of these two parts to enhance robustness in the future.

%
%
\newpage
\bibliographystyle{splncs04}
\bibliography{egbib}
\end{document}